\pdfoutput=1

\documentclass[11pt]{article}

\usepackage[]{acl}

\usepackage{times}
\usepackage{amsmath} \usepackage{amssymb}
\usepackage{latexsym}
\usepackage{graphicx}
\usepackage{multirow}
\usepackage{mathtools}
\usepackage{multicol}
\usepackage{array}
\usepackage{enumerate}
\usepackage{booktabs}
\usepackage{amsfonts,amssymb}

\usepackage[T1]{fontenc}

\usepackage[utf8]{inputenc}

\usepackage{microtype}
\usepackage{amsmath}
\usepackage{autobreak}

\usepackage{inconsolata}
\usepackage{threeparttable}
\usepackage{setspace}
\title{Empowering Backbone Models for Visual Text Generation with Input Granularity Control and Glyph-Aware Training}

\author{Wenbo Li$^{1}$\thanks{~~Work done during internship in Baidu.}, Guohao Li$^{2}$, Zhibin Lan$^{1}$, Xue Xu$^{2}$,Wanru Zhuang$^{1}$, \\ \textbf{Jiachen Liu$^{2}$, Xinyan Xiao$^{2}$, Jinsong Su\textsuperscript{1,3}\thanks{~~Corresponding author.}} \\
    $^1$School of Informatics, Xiamen University, China,\\
    $^2$Baidu Inc., Beijing, China \\ 
    $^3$Shanghai Artificial Intelligence Laboratory
    \\
 \texttt{\small\ liwenbo@stu.xmu.edu.cn}~~ 
 \texttt{\small liguohao@baidu.com}~~
 \texttt{\small jssu@xmu.edu.cn}
 \\
 }

\begin{document}
\maketitle
\begin{abstract}
Diffusion-based text-to-image models have demonstrated impressive achievements in diversity and aesthetics but struggle to generate images with legible visual texts. Existing backbone models have limitations such as misspelling, failing to generate texts, and lack of support for Chinese text, but their development shows promising potential. In this paper, we propose a series of methods, aiming to empower backbone models to generate visual texts in English and Chinese. We first conduct a preliminary study revealing that Byte Pair Encoding (BPE) tokenization and the insufficient learning of cross-attention modules restrict the performance of the backbone models. Based on these observations, we make the following improvements: (1) We design a mixed granularity input strategy to provide more suitable text representations; (2) We propose to augment the conventional training objective with three glyph-aware training losses, which enhance the learning of cross-attention modules and encourage the model to focus on visual texts. Through experiments, we demonstrate that our methods can effectively empower backbone models to generate semantic relevant, aesthetically appealing, and accurate visual text images, while maintaining their fundamental image generation quality. 
\end{abstract}

\section{Introduction}

\begin{figure}[htbp]
    \centering \includegraphics[width=0.48\textwidth]{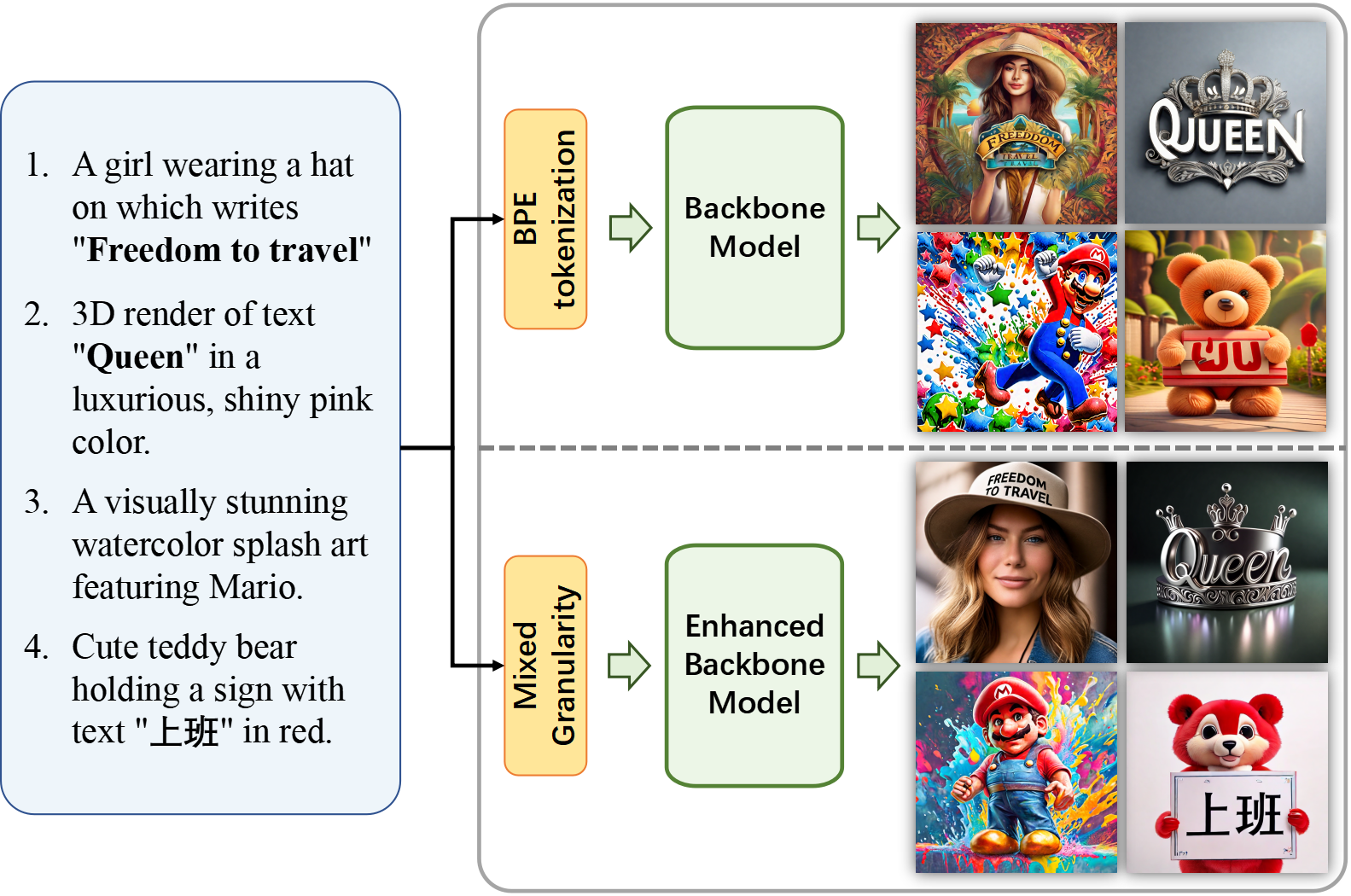} 
    \caption{Comparison between the backbone models (top) and our models (bottom). Our methods can empower the backbone models to generate complex (top left), artistic (top right) visual texts while maintaining fundamental image generation quality (bottom left). Besides, our method can be transferred to Chinese text generation (bottom right).}
    \label{Figure.tech_line}
    \vspace{-0.2cm}
\end{figure}

Recently, diffusion-based models \cite{DBLP:conf/nips/HoJA20, DBLP:conf/cvpr/RombachBLEO22, DBLP:conf/nips/SahariaCSLWDGLA22, DBLP:journals/corr/abs-2211-01324, DBLP:conf/iccv/ZhangRA23, DBLP:journals/corr/abs-2311-17042} have revolutionized the field of text-to-image generation, particularly in terms of diversity and aesthetics. Among various text-to-image tasks, visual text generation has attracted much attention due to the growing demand for generating images containing visual texts in the AI art community and commercial fields. Despite their attractiveness, this task remains challenging, as most current diffusion models struggle to produce images with precise, readable visual texts. At present, dominant studies on this task can be roughly divided into two categories. Some researchers focus on adding additional conditions to reduce the difficulty of generating images with visual texts, resulting in restricted diversity and visual texts not coherent with the background. \cite{DBLP:conf/nips/ChenHL0CW23, DBLP:journals/corr/abs-2303-17870, DBLP:conf/nips/YangGYLDH023, DBLP:journals/corr/abs-2311-03054, DBLP:journals/corr/abs-2311-16465, DBLP:journals/corr/abs-2312-04884}. Other researchers directly explore the performance of backbone models on visual text generation, which avoids the limitations of the previous type of methods but suffers from challenges such as misspelling, ignoring, and repeating words. To deal with these issues, early studies \cite{DBLP:conf/nips/SahariaCSLWDGLA22, DBLP:journals/corr/abs-2211-01324, DBLP:conf/acl/LiuGSCRNBM0C23} explore various text encoders to address misspelling issues. Recent commercial models such as Dall-E 3 \cite{BetkerImprovingIG} and Stable Diffusion 3 \cite{DBLP:journals/corr/abs-2403-03206} demonstrate remarkable performance, further validating the potential of this research direction. However, they lack support for other languages, such as Chinese.

\begin{figure*}[htbp]
    \centering \includegraphics[width=1.0\textwidth]{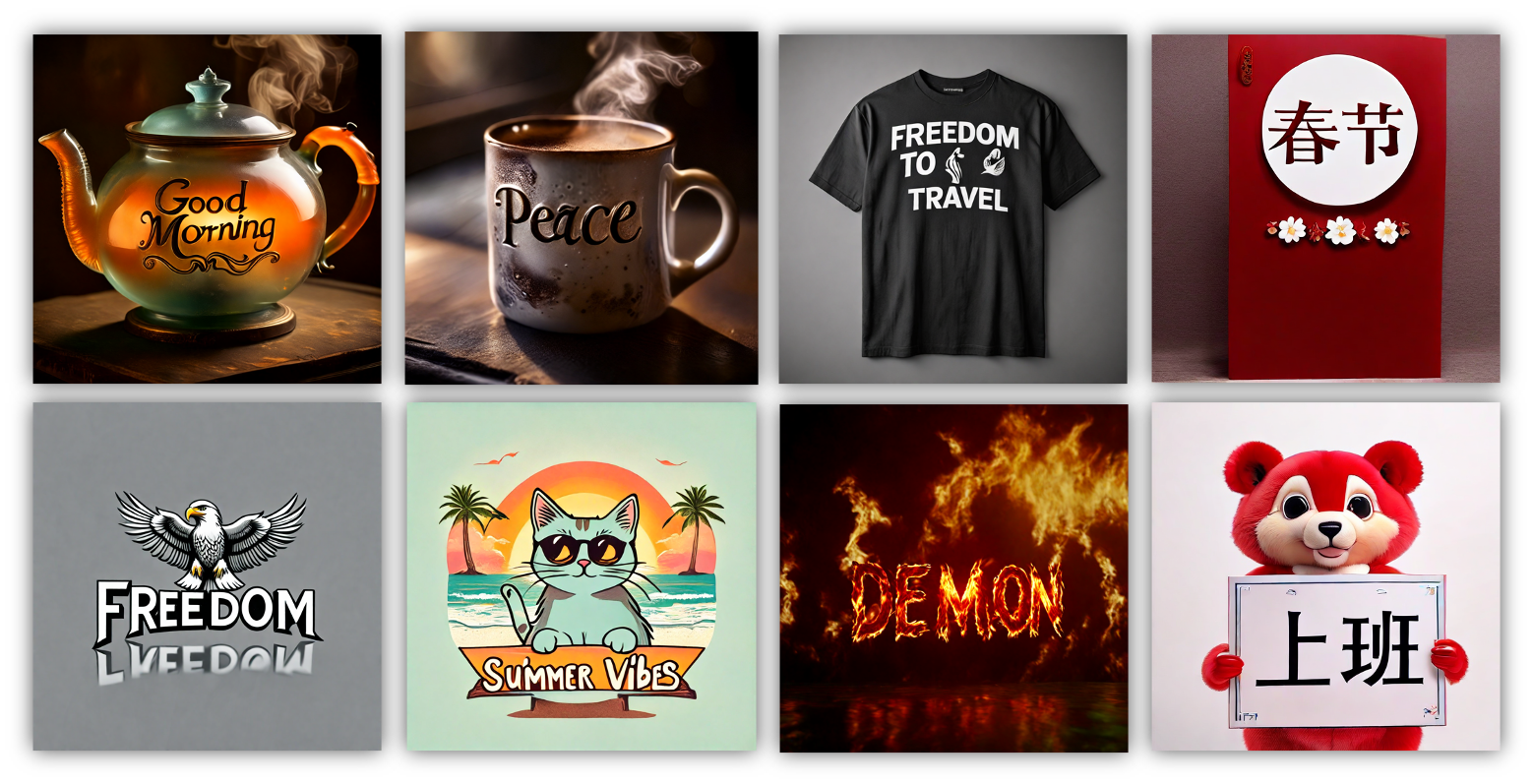} 
    \caption{Visual text generation results of our models. Our methods significantly empower the backbone models to generate semantic relevant, visual appealing visual text images generation in English and Chinese.}
    \label{Figure.showcase}
\end{figure*}

To explore potential avenues for improvements, we first conduct several preliminary experiments and observe that the visual text generation performance of backbone models are mainly constrained for two reasons: First, BPE tokenization requires the model to combine subwords to form complete visual words, increasing the difficulty of generating visual texts. Second, The model is unable to effectively bind visual texts to the corresponding text tokens due to the insufficient learning of the cross-attention modules. 

Based on these analyses, we propose a series of methods that significantly improve the visual text generation capability of backbone models, as shown in Figure \ref{Figure.tech_line}. Specifically, we first introduce a \emph{mixed granularity input} strategy that provides more suitable text representations. Then, we augment the conventional MSE loss with three glyph-aware losses: (1) \emph{attention alignment loss} refines the cross-attention maps, thereby better binding visual texts to their corresponding text tokens; (2) \emph{local MSE loss} highlights the importance of visual text areas; (3) \emph{OCR recognition loss} encourages the model to generate accurate visual texts.

Figure \ref{Figure.showcase} demonstrates that our methods effectively enhance the backbone model's visual text generation ability while maintaining its fundamental capabilities. Particularly, our methods can be transferred to the generation of Chinese texts.


\section{Related Work}
\subsection{Visual Text Generation}
Recent studies on visual text generation primarily focus on introducing additional conditions, such as rendered text images, or position coordinates during inference.

Some works concatenate  representations of the rendered text image with the latent variable as the model input. For example, TextDiffuser \cite{DBLP:conf/nips/ChenHL0CW23} and GlyphDraw \cite{DBLP:journals/corr/abs-2303-17870} concatenate the representation of position-aware mask with the latent variable, and utilize pre-trained models to generate positional information. UDiffText \cite{DBLP:journals/corr/abs-2312-04884} utilizes an inpainting model that considers concatenation of the position mask, the masked image, and the original image as input. Instead of introducing additional conditions through concatenation, some works also explore to utilize auxiliary modules. GlyphControl \cite{DBLP:conf/nips/YangGYLDH023} use a ControlNet, which receives images with rendered texts as input. Building upon this, AnyText \cite{DBLP:journals/corr/abs-2311-03054} introduces a fusion network that receives position and image masks to support more flexible position control and image editing. Apart from these, several works add special tokens representing additional conditions. For example, TextDiffuser-2 \cite{DBLP:journals/corr/abs-2311-16465} adds additional position tokens into the text encoder to generate text based on the predicted coordinates.

However, the above studies still suffer from the following limitations: (1) The use of these conditions constrains the overall composition of the image, causing issues of restricted diversity and visual texts not coherent with backgrounds; (2) Users are required to provide additional conditions, leading to inconvenience in usage.

\subsection{Text-to-Image Backbone Models}
Some researchers focus on enhancing the overall capabilities of text-to-image backbone models. Early works in this regard aim at addressing spelling errors by experimenting with various text encoders. For example, Imagen \cite{DBLP:conf/nips/SahariaCSLWDGLA22} replaces CLIP \cite{DBLP:conf/icml/RadfordKHRGASAM21} with T5 \cite{DBLP:journals/corr/abs-1910-10683}, eDiff-I \cite{DBLP:journals/corr/abs-2211-01324} uses both CLIP and T5.

Additionally, some researchers find that tokenization methods influence the model's ability to generate visual texts. \citet{DBLP:conf/acl/LiuGSCRNBM0C23} believe that the primary reason for spelling errors lies in the lack of character-level glyph information caused by BPE tokenization, and propose to solve this by adopting the character-level text encoder ByT5 \cite{DBLP:journals/corr/abs-2105-13626}.

Recently, some commercial models, such as Dall-E 3 \cite{BetkerImprovingIG} and Stable Diffusion 3 \cite{DBLP:journals/corr/abs-2403-03206} show outstanding performance in visual text generation. This demonstrates that with the development of backbone models, the performance of visual text generation is concurrently improving. However, these commercial models only support English, leaving the generation of visual texts in other languages unsolved.

In this work, we propose a series of methods, which empower the backbone models with the ability to generate accurate and aesthetic visual texts in two aspects. First, we propose a mixed granularity input strategy to provide more suitable text representations.
Second, we augment the conventional training objective with three glyph-aware losses.

\begin{figure}[!t]
    \centering \includegraphics[width=0.4\textwidth]{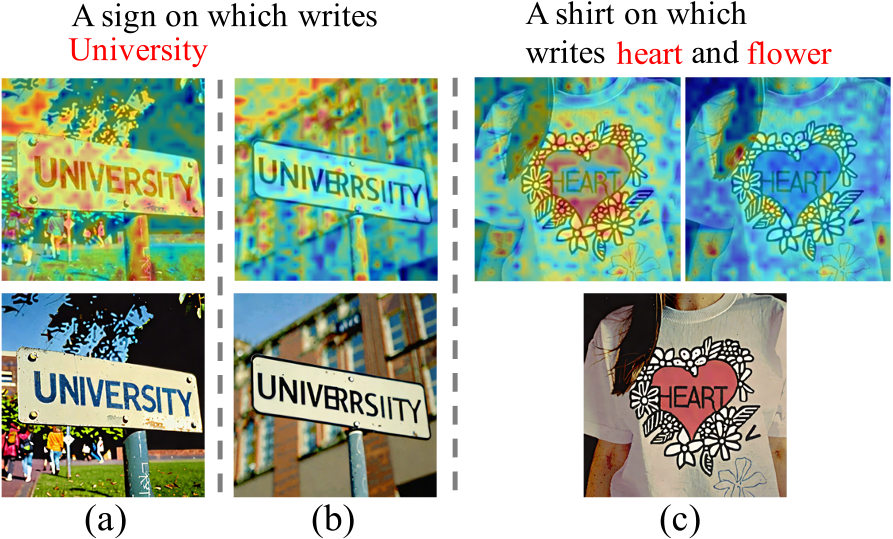}
    \caption{Visualization of the cross-attention maps. (a): ``\textit{University}'' is correctly spelled, the token has large values on the corresponding areas. (b): ``\textit{University}'' is not correctly spelled, token ``\textit{university</w>}'' fails to focus on the corresponding area. (c): The token ``\textit{heart</w>}`` attends to the corresponding area, thus is correctly generated, while the token ``\textit{flower</w>}``  highlights irrelevant region and fails to generate the corresponding visual text.}
    \label{Figure.prestudy}
    \vspace{-0.2cm}
\end{figure}

\section{Preliminary Study}

In this section, we first introduce the basic concepts of the diffusion based text-to-image backbone model, and then conduct experiments to identify potential avenues for improvements.
 
\subsection{Diffusion Based Text-to-Image Backbone Models}

\textbf{Model Architecture.} \quad The commonly-used architecture of text-to-image backbone models derives from the latent diffusion model \cite{DBLP:conf/cvpr/RombachBLEO22}, which is composed of three modules: (1) a VAE \cite{DBLP:journals/corr/KingmaW13} consists of an encoder to compress images into the latent space, and a decoder to reverse them back; (2) a UNet \cite{DBLP:journals/corr/RonnebergerFB15} denoiser \(\epsilon_\theta\) performs diffusion denoising process at latent space; (3) a text encoder \(\mathcal{T}\) encodes the text prompt into representation \(\boldsymbol{c}\).

\noindent \textbf{Diffusion Process.} \quad
This process defines a Markov chain of forward diffusion process which continually applies the noise sampled from a Gaussian distribution to the real data \(\boldsymbol{z}_0 = \mathcal{E}(\boldsymbol{x}_0)\):
\vspace{-0.1cm}
\begin{equation}
\vspace{-0.1cm}
q(\boldsymbol{z}_t|\boldsymbol{z}_{t-1}) := \mathcal{N}(\boldsymbol{z}_t; \sqrt{{\alpha}_t} \boldsymbol{z}_{t-1}, (1 - {\alpha}_t)\boldsymbol{I}), \label{equ:1}
\end{equation}
where \({\alpha}_t\) is a time-aware schedule. As \(t\) increases, \(\boldsymbol{z}_t\) asymptotically approaches the noise in a standard Gaussian distribution. 

The UNet denoiser \(\boldsymbol{\epsilon}_\theta\) is trained to predict the noise \(\boldsymbol{\epsilon}_t\) added to the image at timestep \(t\), thereby reversing the Markov chain. A mean squared error (MSE) loss is utilized to supervise the training:
\begin{equation} 
\mathcal{L}_{mse} = \mathbb{E}_{\boldsymbol{z}_0, \boldsymbol{c}, \boldsymbol{\epsilon}_t, t} \left[ || {\boldsymbol{\epsilon}}_{\theta}(\boldsymbol{z}_0, t, \boldsymbol{c}) - \boldsymbol{\epsilon}_t ||_2^2 \right]. \label{equ:4}
\end{equation}

To add conditional guidance, the representation \(\boldsymbol{c}\) is fed into each cross-attention block of the UNet model as:
\begin{equation}
    \begin{aligned}
        Attn(\boldsymbol{z}_t, \boldsymbol{c}) = \text{Softmax}(\frac{Q(\boldsymbol{z}_t) \cdot K(\boldsymbol{c})^T}{\sqrt{d}})V(\boldsymbol{c}),
    \end{aligned}
\end{equation}
where \(Q\), \(K\) and \(V\) denote the query, key and value projections, and \(d\) denotes the output dimension.


\begin{figure*}[!t] 
    \centering \includegraphics[width=0.95\textwidth]{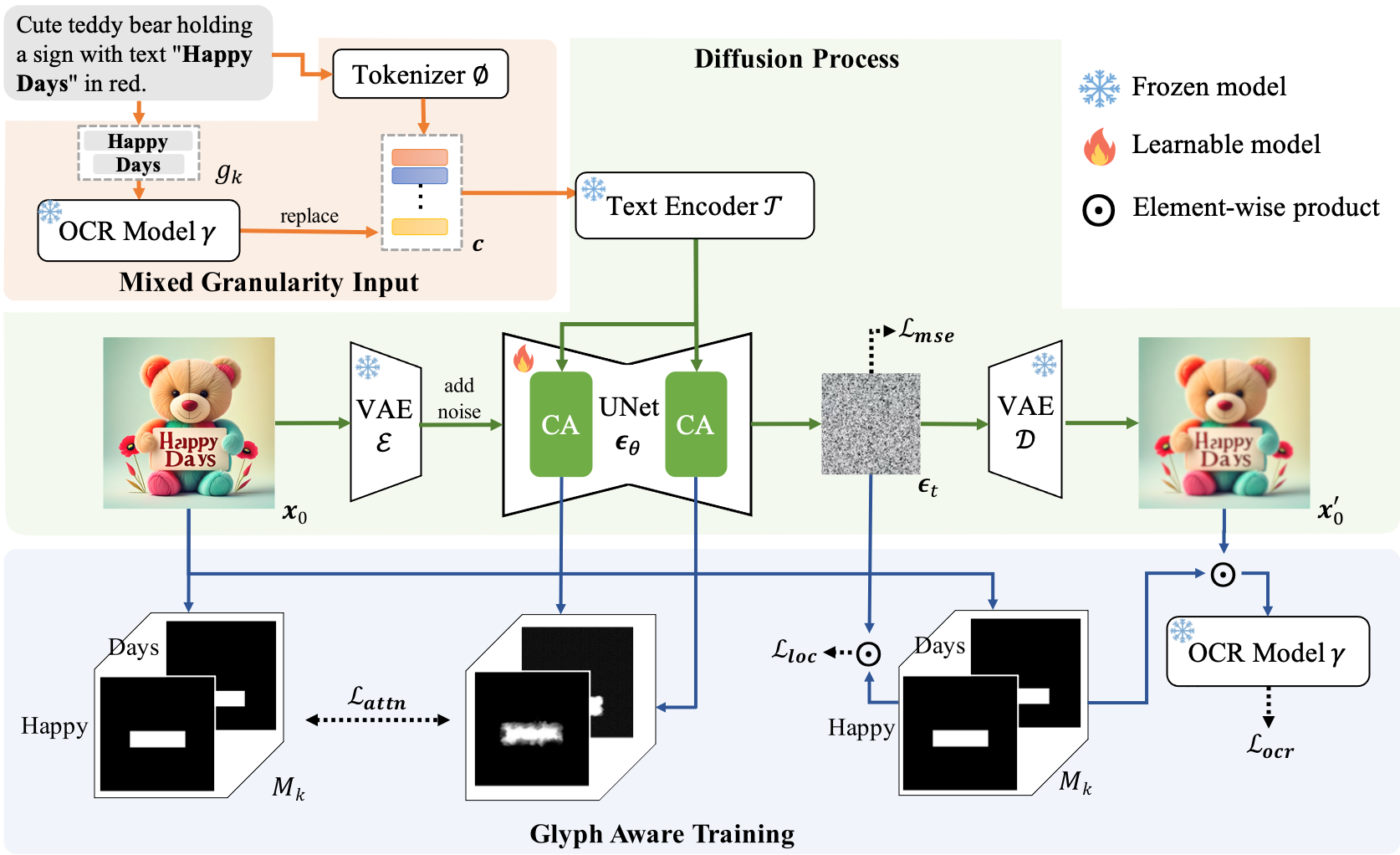} 
    \caption{The framework of our methods. The Mixed Granularity Input strategy considers glyph words as whole units to provide more suitable text representations. The Glyph Aware Training includes three losses: (1) the {attention alignment loss} enhances the learning of cross-attention modules; (2) the {local MSE loss} highlights the importance of visual text areas; (3) the {OCR recognition loss} encourages the model to generate accurate visual texts.}
    \label{Figure.model}
\end{figure*}

\subsection{Experimental Analyses}
To identify avenues for improvements, we use the commonly-used backbone model--SD-XL \cite{DBLP:journals/corr/abs-2307-01952} to conduct two groups of experiments.

In the first group of experiments, we investigate the effect of BPE tokenization on two subsets: (1)
\(S_{1}\), where words are split into subwords by BPE tokenization, and (2) \(S_{2}\), consisting of words that remains the same after BPE tokenization. To eliminate the impact of word frequency and length, we select 100 words for each subset from 5,000 common words with lengths ranging from 5 to 8 letters\footnote{\url{https://github.com/first20hours/google-10000-english}}. Results show that the model achieves an accuracy of 0.3 in \(S_{1}\), compared to 0.46 in \(S_{2}\), indicating that \textbf{BPE tokenization increases the difficulty for the model in generating visual texts}, as it splits a word into subwords and requires the model to combine them into a complete visual word.

As stated in previous works \cite{DBLP:conf/iclr/HertzMTAPC23, DBLP:journals/tog/CheferAVWC23}, the cross-attention maps of the UNet can reflect the relevance between generated objects and corresponding text tokens. Similarly, visual texts can also be treated as objects, and texts to be generated, which we refer to as \textit{glyph texts}, should therefore have a robust relationship with the corresponding visual texts in the image. In the second group of experiments, we extract and visualize the cross-attention maps for glyph tokens at the last timestep, as depicted in Figure \ref{Figure.prestudy}. We can clearly observe that cross-attention maps with corresponding visual texts generated are correctly localized, while the maps that do not have corresponding visual texts generated highlights irrelevant regions. Thus, we conclude that \textbf{glyph tokens indeed have a robust relationship with visual text areas through cross-attention mechanism}, which the model fails to effectively capture.

In summary, based on our experimental analyses, we believe that BPE tokenization and the insufficient learning of cross-attention modules constrain the model's ability to correctly generate visual texts.

\section{Methods}
Based on the observations from our preliminary study, we propose a series of methods to improve the visual text generation capability of backbone models. As shown in Figure \ref{Figure.model}, our improvements mainly involve two aspects:
(1) we introduce a mixed granularity input strategy to replace the BPE subword input;
(2) we augment the conventional training objective with three glyph-aware training losses, which regulates the cross-attention maps and encourage the model to focus on visual texts.

\subsection{Mixed Granularity Input}
\begin{figure}[t]
    \centering \includegraphics[width=0.45\textwidth]{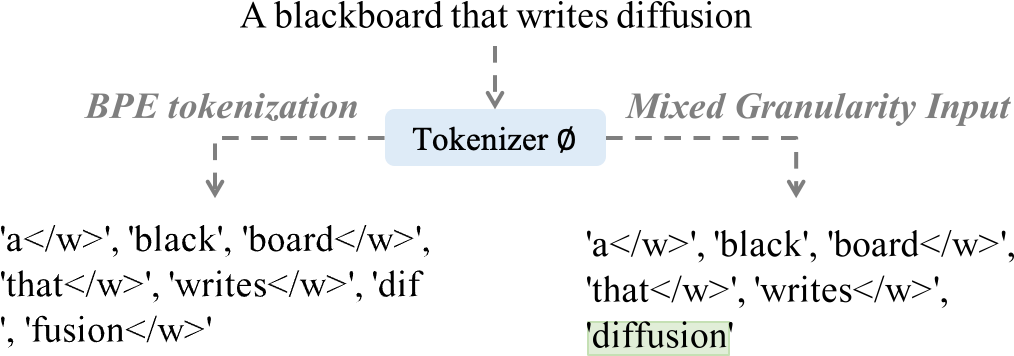} 
    \caption{Mixed granularity input. The word ``\textit{diffusion}'' is considered as a whole instead of being tokenized.}
    \label{Figure.mix_granularity}
    \vspace{-0.3cm}
\end{figure}

Our preliminary study reveals that BPE tokenization constrains the performance of the model, highlighting the necessity to represent glyph texts in a more suitable granularity. In this regard, previous studies \cite{DBLP:conf/acl/LiuGSCRNBM0C23, DBLP:journals/corr/abs-2311-16465,DBLP:journals/corr/abs-2312-04884} commonly utilize character-level tokenization, which splits words into characters. However, as stated in our preliminary study, this split challenges the model to combine characters into a complete visual word. To deal with this issue, we consider each glyph word as a whole within the model, as shown in Figure \ref{Figure.mix_granularity}. Given the impracticality of including every word in the vocabulary, a method is needed to get the embedding for every word. Therefore, we extract intermediate features from the OCR model as new text embeddings following \citet{DBLP:journals/corr/abs-2311-03054}, which inherently possess sufficient glyph information. Specifically, for a user prompt \(\boldsymbol{y}\) containing \(N\) glyph words \(g_1, g_2, \dots, g_N\), we render each glyph word into an image without providing positional information, resulting in an image sequence \(\boldsymbol{I_g}\). Then, we feed them into the OCR model \(\gamma\), where the text embedding \(\boldsymbol{c}\) is refined as follows:
\begin{equation}
\boldsymbol{c} = \mathcal{T}(\phi(\boldsymbol{y}), \mathcal{\xi}(\gamma(\boldsymbol{I_g}))),
\end{equation}
where \(\mathcal{T}\) is the CLIP text encoder, \(\phi\) is the BPE tokenizer, and \(\xi\) is a linear module.

\subsection{Glyph-Aware Training}
Formally, the overall training objective can be formulated as:
\begin{equation}
\begin{aligned}
    \mathcal{L} &= \mathcal{L}_{mse} + \lambda_{1} \cdot \mathcal{L}_{attn} + \lambda_{2} \cdot \mathcal{L}_{loc} \\ &+ (1-\lambda_{1}-\lambda_{2})\cdot \mathcal{L}_{ocr},
\label{equ:6}
\end{aligned}
\end{equation}
where \(\mathcal{L}_{attn}\), \(\mathcal{L}_{loc}\) and  \(\mathcal{L}_{ocr}\) denote the attention alignment loss, the local MSE loss, and the OCR recognition loss, respectively.

\subsubsection{Attention Alignment Loss \(\mathcal{L}_{attn}\)}
To enhance the learning of cross-attention modules, we introduce an attention alignment loss, which encourages the model to ensure that each visual text mainly attends to the corresponding glyph token. Specifically, the cross-attention map between the intermediate feature of the noisy latent variable \(\boldsymbol{z}_t\) and the refined representation \(\boldsymbol{c}_g\) of glyph tokens can be calculated as:
\begin{equation}
    \begin{aligned}
        CA(\boldsymbol{z}_t, \boldsymbol{c_g}) = \text{Softmax}(\frac{Q(\boldsymbol{z}_t) \cdot K(\boldsymbol{c_g})^T}{\sqrt{d}}).
    \end{aligned}
\end{equation}

To encourage that each visual text has large values in the corresponding area, we minimize the distance between the cross-attention maps and the corresponding segmentation masks of visual texts, which is defined as follows:
\vspace{-0.1cm}
\begin{equation}
    \begin{aligned}
        \mathcal{L}_{attn} = \frac{1}{N} \sum_{k=1}^N \left\| CA(\boldsymbol{z}_t, \boldsymbol{c}_g^k) - M_k \right\|_2^2,
    \end{aligned}
\end{equation}
where \(M\) denotes the segmentation mask of the \(k\)-th visual text corresponding to its glyph token.

Through this training process, the model can effectively capture a more robust understanding of the relationships between the visual texts and glyph tokens, thus faithfully generating the desired visual texts.


\begin{table*}[]
\centering
\begin{small}
\renewcommand\arraystretch{1.5}
\resizebox{\textwidth}{!}{\begin{tabular*}{\linewidth}{@{}lccccccc@{}} 
\toprule[1pt]
{\textbf{Metrics}} & 
\textbf{DeepFloyd} & 
\textbf{SD-XL} & 
\textbf{SDXL-Turbo} &
\textbf{LCM-LoRA} &
\textbf{SD Cascade} &
\textbf{Ours(XL)} &
\textbf{Ours(Turbo)}
\\ \hline

\multicolumn{8}{c}{\textit{\textbf{Quantitative Results }}} \\ \hline
{{CLIP score}} & 28.256 & 28.056 & 30.001 & 26.631  & \textbf{30.102} & 29.168 & 29.394 \\ 
OCR (Precision) & 0.153 & 0.116 & 0.315 & 0.032 & 0.298 & \textbf{0.348} & \textbf{0.360} \\
OCR (Recall) & 0.210 & 0.138 & \textbf{0.337} & 0.038 & 0.308 & 0.283 & 0.307 \\
OCR (F1 score) & 0.177 & 0.125 & 0.326 & 0.035 & 0.303 & 0.312 & \textbf{0.331} \\ \hline
\multicolumn{8}{c}{\textit{\textbf{User Study Results}}} \\ \hline
{Text Aesthetics} & 0.100 & 0.050 & 0.100 & 0.016 & 0.200 & \textbf{0.259} & \textbf{0.275} \\ 
{Text Accuracy} & 0.131 & 0.041 & 0.134 & 0.006 & 0.188 & \textbf{0.234} & \textbf{0.266} \\ 
{Semantic Relevance} & 0.066 & 0.022 & 0.059 & 0.009 & 0.178 & \textbf{0.303} & \textbf{0.363} \\
{Image Aesthetics} & 0.038 & 0.034 & 0.053 & 0.031 & 0.178 & \textbf{0.287} & \textbf{0.378} \\
\bottomrule[1pt]
\end{tabular*}}
\caption{Quantitative results of English text generation compared with other backbone models. `XL', `Turbo' denotes SD-XL, SDXL-Turbo. Our models achieve the best results in terms of most metrics.}
\label{figures:MainExp}
\end{small}
\vspace{-0.1cm}
\end{table*}

\subsubsection{Local MSE Loss \(\mathcal{L}_{loc}\)}

Since the MSE loss only measures pixel-wise distance and lacks additional focus on visual text areas, we apply a weighting strategy to the MSE loss following \citet{DBLP:journals/corr/abs-2303-17870}, which we refer to as the \emph{local MSE loss}. To mitigate the impact of visual text area size, we add a weighting term \({w}\) which is the ratio of the image area to the visual text area. Formally, the local MSE loss can be formulated as:
\vspace{-0.2cm}
\begin{equation}
\vspace{-0.2cm}
\begin{aligned}
    \mathcal{L}_{loc} &= \frac{1}{N} \sum_{k=1}^N w^k \cdot\mathcal{L}_{loc}^k, \\ 
    \mathcal{L}_{loc}^k &= \mathbb{E}_{\boldsymbol{z}_0, \boldsymbol{\epsilon}_t, t} \left[M_k \odot \left\| {\boldsymbol{\epsilon}}_{\theta}(\boldsymbol{z}_0, t, \boldsymbol{c}) - \boldsymbol{\epsilon}_t \right\|_2^2 \right].
\end{aligned}
\end{equation}

\subsection{OCR Recognition Loss \(\mathcal{L}_{ocr}\)}
To further encourage the model to generate accurate visual texts, we introduce an OCR recognition task. At each training step, we can estimate the fully denoised image latent variable \({\boldsymbol{z}}_0'\), as implemented in DDPM \cite{DBLP:conf/nips/HoJA20}. We then input this latent variable into the VAE decoder to obtain an approximate image \({\boldsymbol{x}}_0'\), which is subsequently fed into the OCR model for recognition. As implemented in the training of the OCR model, we use the CTC loss \cite{DBLP:conf/icml/GravesFGS06} to refine the predicted results. Since this estimation introduces more distortion as \(t\) increases, we add a weighting term related to \(t\), which is set as \(\bar{\alpha}_t\) following \citet{DBLP:journals/corr/abs-2311-03054}. The OCR recognition loss can be formulated as:
\vspace{-0.2cm}
\begin{equation}
\vspace{-0.2cm}
    \mathcal{L}_{ocr} = \frac{1}{N} \sum_{k=1}^N \bar{\alpha}_t \cdot CTC({\boldsymbol{x}}_0' \odot M_k, g_k),
\end{equation}
where \(CTC(\cdot)\) denotes the CTC loss function.

\section{Experiments}
\subsection{Dataset}
To better unleash the potential of the model, we require a large-scale, high-quality dataset that satisfies the following criteria:
\vspace{-\topsep}
\begin{itemize}
\setlength{\itemsep}{0.1cm}
\setlength{\parsep}{2pt}
\setlength{\parskip}{2pt}
    \item The dataset should contain images with clear and recognizable visual texts.
    \item The visual texts in the images should occupy a prominent area and be coherent with the background.
    \item The captions should include detailed descriptions of the visual texts.
    \item The aesthetic quality of the images should be comparable to those used for pre-training.
\end{itemize}

Following the aforementioned criteria, we construct an English dataset consisting of 240K samples by filtering internal datasets, and a Chinese dataset containing 50K synthetic samples using image-to-image models and rendering tools\footnote{\url{https://pypi.org/project/pillow/}}. More details are introduced in Appendix \ref{sec:appendix_A}.

\subsection{Experimental Setting}
\noindent \textbf{Evaluation Metrics.} \quad We quantify the visual text generation quality from two aspects: (1) \textbf{CLIP score} \cite{DBLP:conf/emnlp/HesselHFBC21} measures the semantic relevance between the generated image and the input prompt by calculating the cosine similarity of their representations from CLIP image and text model \cite{DBLP:conf/icml/RadfordKHRGASAM21}. (2) \textbf{OCR Accuracy} detects the texts in the generated images utilizing OCR tools. We calculate the precision, recall and F1 score between the detected texts and the ground truths. Furthermore, we evaluate the model 's fundamental capability through \textbf{FID} \cite{DBLP:conf/nips/HeuselRUNH17} score, which compares the distribution of generated images with that of real images. Note that we are unable to calculate the FID score in our main experiments, due to the lack of source images.

\begin{figure*}[t]
    \centering \includegraphics[width=0.95\textwidth]{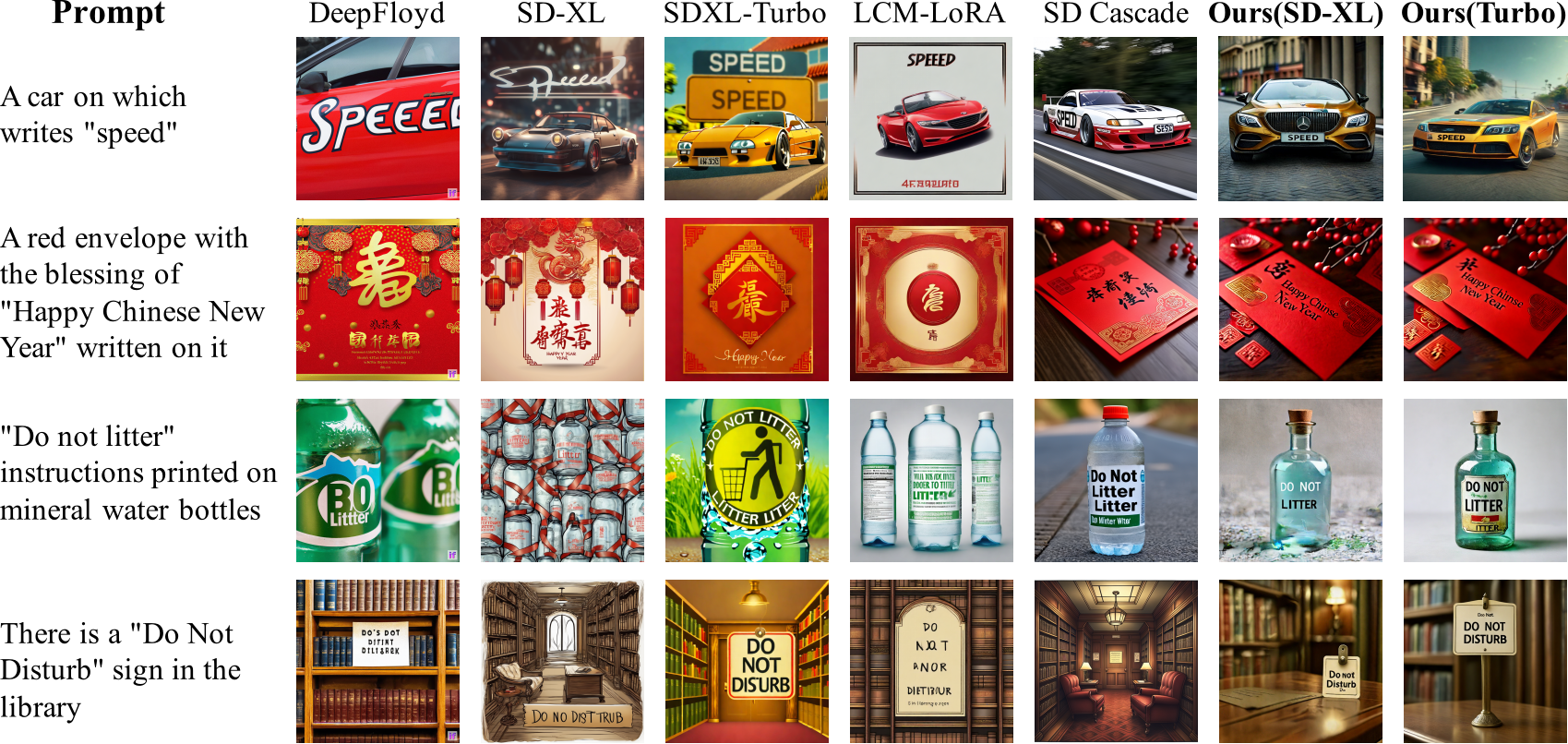} 
    \caption{Visualization of generating English texts compared with other backbone models.}
    \label{Figure.main_vis_en}
    \vspace{0.2cm}
\end{figure*}

\noindent \textbf{Implementation Details.}
\quad We train our models based on SDXL-base-1.0 and SDXL-Turbo. We utilize the PaddleOCR v4 model\footnote{\url{https://github.com/PaddlePaddle/PaddleOCR}} to extract intermediate features, perform the OCR recognition task, and conduct evaluation. We set \(\lambda_{1}\) and \(\lambda_{2}\), to 0.4, 0.2, respectively, determined by a grid search on the validation set, which are varied from 0.1 to 1.0 with an interval of 0.1. We set the learning rate to 2e-5 and conduct a total of 10K steps of training. The overall training process takes 7 hours and 50 minutes on 8 A800 GPUs.

\subsection{Quantitative Results}
As shown in Table \ref{figures:MainExp}, we conduct quantitative comparison with existing backbone models on the ChineseDrawText benchmark \cite{DBLP:journals/corr/abs-2303-17870}. We compare our models with DeepFloyd \cite{DeepFloyd}, SD-XL, SDXL-Turbo \cite{DBLP:journals/corr/abs-2311-17042}, LCM-LoRA \cite{DBLP:journals/corr/abs-2311-05556}, and SD-Cascade \cite{pernias2023wuerstchen}. Results show that our models outperform other baseline models under the majority of metrics.

\subsection{User Study}
To further validate the effectiveness of our proposed methods, we conduct a human evaluation comparing our English models with other baseline models on ChineseDrawText benchmark. Three raters are asked to compare these images from four dimensions including text aesthetics, text accuracy, semantic relevance, and image aesthetics, and then select the images they prefer. Throughout the process, all raters are unaware of which model the image is generated from. The results in Table \ref{figures:MainExp} show that human raters greatly prefer our models on all aspects, which further validates the effectiveness of our approaches in generating high-quality and visual text images. The detailed participant instruction are listed in Appendix \ref{sec:appendix_B}.

\begin{figure}[ht]
    \centering \includegraphics[width=0.45\textwidth]{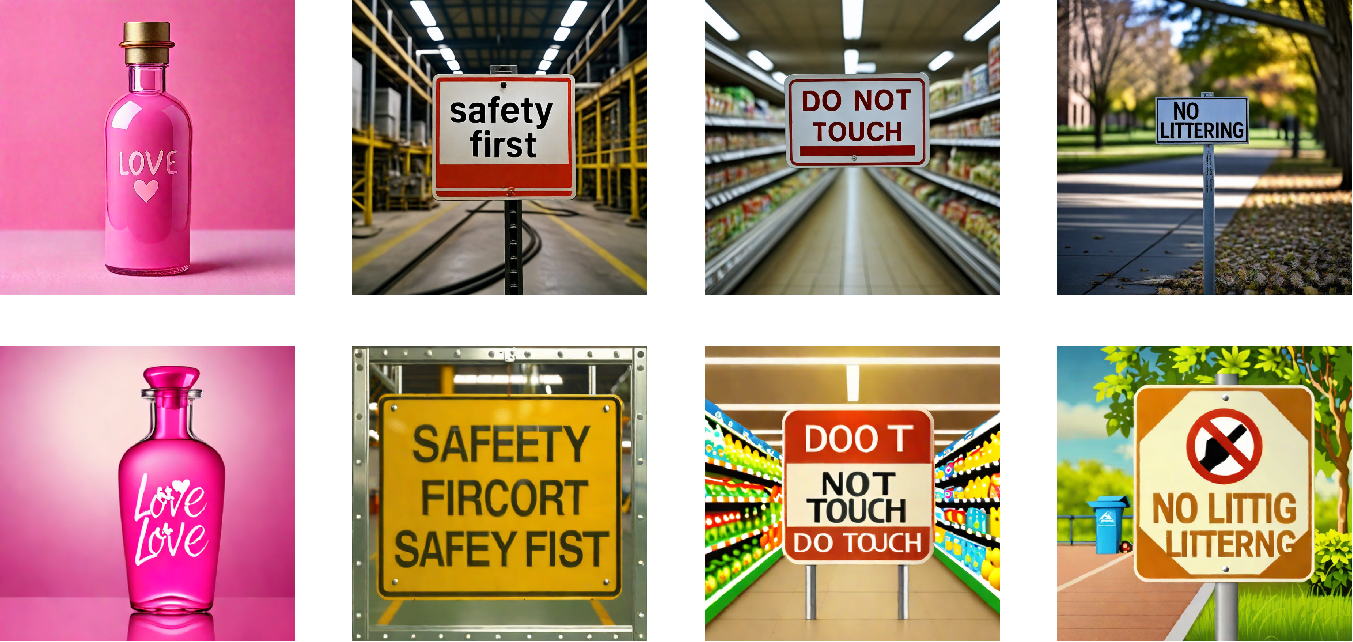} 
    \caption{Comparison between Ours(Turbo) at the top and SDXL-Turbo at the bottom.}
    \label{Figure.fail}
\end{figure}

\subsection{Qualitative Results}
To provide more straightforward comparison, we provide some visualization samples from the test set in Figure \ref{Figure.main_vis_en}. We can clearly observe that our models better capture the semantic relevance, thus generate visual texts at the reasonable place. For example, our models generate visual text ``\textit{Speed}'' at the front of the car (line 1), while some baselines (SD-XL, SDXL-Turbo, LCM-LoRA) fail to capture the guidance ``\textit{write on the car}''. Besides, as shown in line 3, our models effectively avoid issues like misspelling (Deepfloyd, SDXL-Turbo, LCM-LoRA), repeating (SD Cascade, SDXL-Turbo) and ignoring words (Deepfloyd), or failing to understand the instruction (SD-XL). We provide visualization comparison and more showcases in Appendix \ref{sec:appendix_C}.

Regarding the failure cases, we analyze why the recall of our model is lower than that of the SDXL-Turbo model. As shown in Figure \ref{Figure.fail}, the SDXL-Turbo model tends to generate repeated words. For example, in the first column, the SDXL-Turbo model generates ``\textit{do}'' and ``\textit{not}'' twice and spells them correctly once, leading to a higher recall. However, this repetition fails to align with human reference, resulting in a lower precision score.

\subsection{Comparison of Generating Images without Visual Texts}

\begin{figure}[t]
    \centering \includegraphics[width=0.48\textwidth]{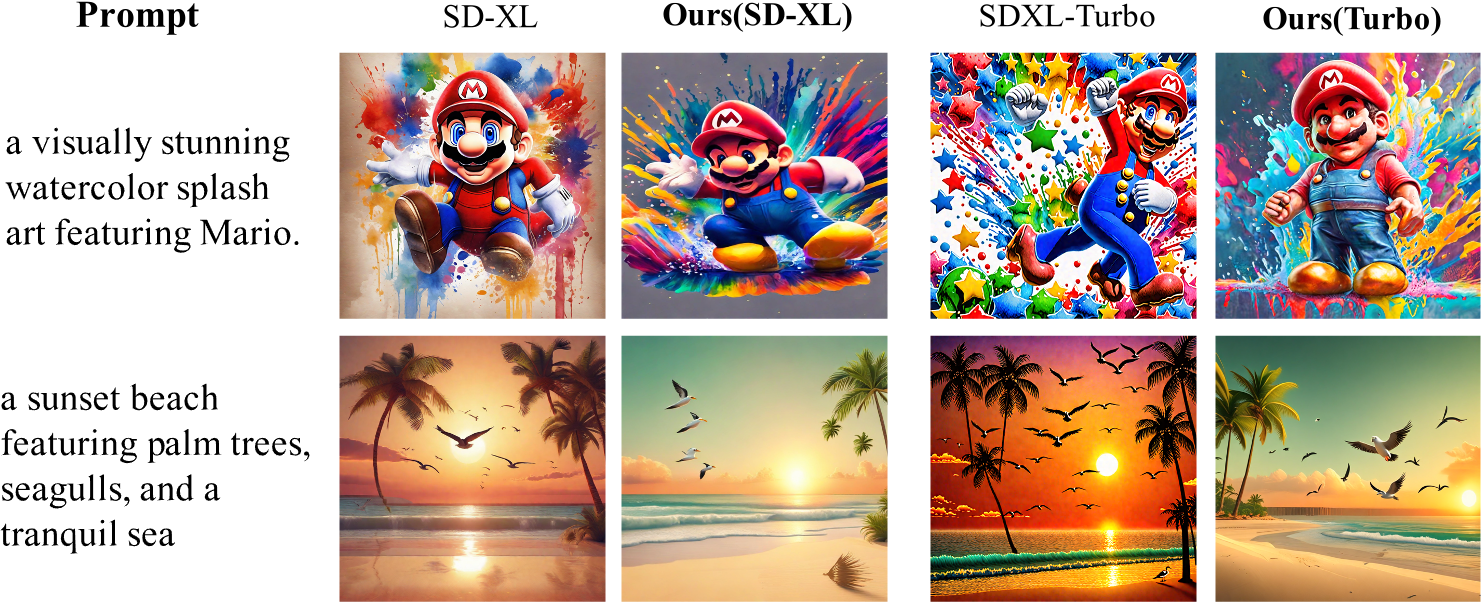}
    \caption{Visualization of generating images without visual texts.}
    \label{Figure.compare_wo_text}
\end{figure}

In order to evaluate the fundamental image generation performance of our models, we use FID to quantify the image quality without visual texts on 5K samples from COCO2017 \cite{DBLP:conf/eccv/LinMBHPRDZ14}, as shown in Table \ref{figures:MSCOCO}. Furthermore, we visualize some image generation examples in Figure \ref{Figure.compare_wo_text}. The quantitative and qualitative results indicate that our models maintain the fundamental capability to generate visual appealing and semantic relevant images.


\begin{table}[]
\centering
\scalebox{0.8}{
\renewcommand\arraystretch{1.5}
\begin{tabular*}{\linewidth}{@{}lc} 
\toprule[1pt]
\makebox[0.35\textwidth][l]{\textbf{Method}} & \makebox[0.1\textwidth][c]{\textbf{FID}}  \\ \hline

Ours(SD-XL) & \textbf{53.4}  \\
Ours(SDXL-Turbo) & \textbf{52.9} \\
SD-XL \cite{DBLP:journals/corr/abs-2307-01952} & 73.6 \\
SDXL-Turbo \cite{DBLP:journals/corr/abs-2311-17042} & 61.9 \\ 
\bottomrule[1pt]
\end{tabular*}}
\caption{COCO zero-shot \(\text{FID}_{\text{5k}}\)(FID) comparison.}
\label{figures:MSCOCO}
\vspace{0.2cm}
\end{table}

\subsection{Ablation Study}
To investigate the effectiveness of each design, we further compare
our SD-XL model with the following variants in Table \ref{figures:Ablation}:

(1) \textit{\(\Rightarrow \)Char+BPE tokenization, and\(\Rightarrow\)BPE tokenization}.\quad In the first variant, we replace our mixed granularity input with the mixture of character-level and BPE tokenization. In the second variant, we only utilize BPE tokenization. As shown in line 2, our mixed granularity input strategy outperforms the mixture of character-level and BPE tokenization. We hypothesize that this is because the model struggles to combine the glyphs of characters to form a complete visual word. The result in line 3 shows that the mixture of character-level and BPE tokenization achieves better results comparing to BPE tokenization, which demonstrate the effectiveness of providing character-level glyph information. 

(2) \textit{w/o \(\mathcal{L}_{attn}\)}. \quad  In this variant, the attention alignment loss \(\mathcal{L}_{ocr}\) is removed. The result in line 4 shows a significant performance drop, which confirms our previous assumption that the insufficient learning of cross-attention modules constrains the visual text generation capability of backbone models.


\begin{table}[]
\centering
\begin{small}
\setlength{\tabcolsep}{2.5mm}
\renewcommand\arraystretch{1.5}
\begin{tabular*}{\linewidth}{@{}lccc@{}} 
\toprule[1pt]
{\textbf{Method}} &  {\textbf{{Pre.}}} & \textbf{{Rec.}} & \textbf{{F1.}} \\ \hline

 Ours & \textbf{0.348} & \textbf{0.283} & \textbf{0.312} \\
\hspace{1em} \(\Rightarrow\) Char+BPE tokenization & 0.231 & 0.193 & 0.210 \\
\hspace{1em} \(\Rightarrow\) BPE tokenization & 0.227 & 0.182 & 0.202 \\
\hspace{1em}w/o \(\mathcal{L}_{attn}\) & 0.278 & 0.236 & 0.250 \\
\hspace{1em}w/o \(\mathcal{L}_{loc}\) & 0.254 & 0.236 & 0.244  \\
\hspace{1em}w/o \(\mathcal{L}_{ocr}\) & 0.270 & 0.218 & 0.241 \\ 
\bottomrule[1pt]
\end{tabular*}
\caption{Ablation on the generation of English texts. \(\Rightarrow\)* means replacing the input granularity with *. \(\mathcal{L}_{loc}\), \(\mathcal{L}_{attn}\) and  \(\mathcal{L}_{ocr}\) denote the local MSE loss, the attention alignment loss, and the OCR recognition loss, respectively. `Pre.' and `Rec.' denote Precision and Recall respectively.}
\label{figures:Ablation}
\end{small}
\vspace{0.2cm}
\end{table}

(3) \textit{w/o \(\mathcal{L}_{loc}\)}.\quad We remove the local MSE loss \(\mathcal{L}_{loc}\) from this variant. The result in line 5 demonstrates a significantly decline in OCR accuracy, indicating that focusing on visual text areas is indeed helpful for generating correctly spelled visual texts.

(4) \textit{w/o \(\mathcal{L}_{ocr}\)}. \quad We remove the OCR recognition loss \(\mathcal{L}_{ocr}\), and observe from line 6 that the performance suffers from a great decline, demonstrating the effectiveness of OCR recognition loss. 

\begin{figure}[t]
    \centering \includegraphics[width=0.45\textwidth]{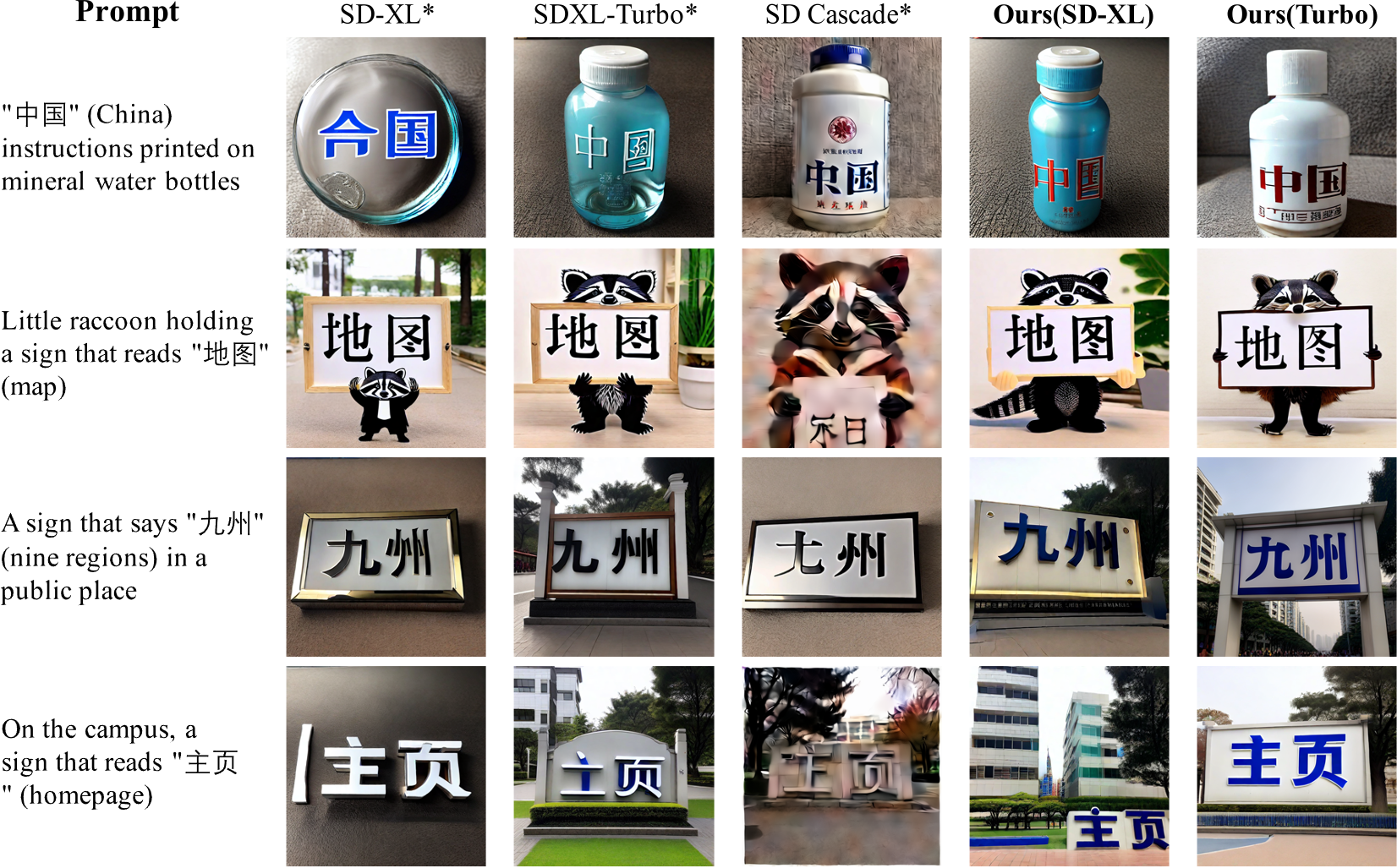} 
    \caption{Visualization of generating Chinese texts compared with other backbone models. * denote the baseline model that is trained on our Chinese dataset for 10K steps.}
    \label{Figure.main_vis_zh}
\end{figure}

\subsection{Chinese Text Generation}
We further explore the effectiveness of our methods to generate Chinese visual texts. 
Instead of considering words as whole units, we use the mixture of character-level and BPE tokenization for Chinese texts due to two reasons. First, Chinese glyphs are excessively complex, resulting in the intermediate features of each character being too similar in distribution to be effectively distinguished. Second, fewer characters are included in each Chinese word, thus is easier to be combined into a complete visual word.

Note that due to the lack of open source Chinese backbone text-to-image models for comparison, we train both our models and the baseline models on our Chinese dataset for 10K steps. We choose SD-XL, SDXL-Turbo and SD Cascade, which achieve relatively better performance in English, as baseline models, and use the prompt templates from the ChineseDrawText benchmark with texts included in our Chinese dataset as test set. 
Quantitative results in Table \ref{figures:MainExpZH} show that our models greatly outperform other baseline models. 
As for qualitative comparison, we visualize some samples from our test set, as shown in Figure \ref{Figure.main_vis_zh}. Our model generates accurate visual texts, while other baseline models fails to correctly generate Chinese texts, indicating that our methods enhance the learning of Chinese texts.
We provide the ablation study for Chinese text generation in Appendix \ref{sec:appendix_D}.


\begin{table}[]
\centering
\begin{small}
\setlength{\tabcolsep}{2.9mm}
\renewcommand\arraystretch{1.5}
{\begin{tabular*}{\linewidth}{@{}lcccc@{}} 
\toprule[1pt]
{\textbf{Method}} & {\textbf{\makebox[0.075\textwidth]{CLIP score}}} &  {\textbf{{Pre.}}} & \textbf{{Rec.}} & \textbf{{F1}} \\ \hline
 


SD-XL* & 29.569 & 0.299 & 0.319 & 0.308 \\
SDXL-Turbo* & 29.827 & 0.275 & 0.295 & 0.284  \\
SD Cascade* & 29.734 & 0.146 & 0.154 & 0.150 \\
Ours(SD-XL) & {29.258} & {0.582} & {0.608} & {0.595} \\

Ours(Turbo) & \textbf{29.919} & \textbf{0.598} & \textbf{0.613} & \textbf{0.606} \\
\bottomrule[1pt]
\end{tabular*}}
\caption{Quantitative results of Chinese text generation. * denotes that these baselines are trained on our Chinese dataset for 10K steps before comparison.}
\label{figures:MainExpZH}
\end{small}
\end{table}

\section{Conclusion}
In this paper, we conduct a preliminary study and find that BPE tokenization, as well as the model's insufficient learning of cross-attention modules, constrains the visual text generation performance of diffusion-based backbone models. Based on these insights, we propose a series of methods, aiming to empower the backbone model with the ability to generate accurate and aesthetically appealing visual text images, while maintaining fundamental image generation quality. Specifically, we introduce a mixed granularity input strategy to provide more suitable text representations. Besides, we augment the conventional training objective with three glyph-aware training losses, which enhance the learning of the cross-attention modules and encourage the model to focus on visual texts. Experiments demonstrate the effectiveness of our methods. Typically, our methods can be transferred to Chinese text generation.

In the future, we intend to explore visual text generation for more languages, and generate texts in different styles \cite{DBLP:journals/corr/abs-2307-02971}. Besides, we also plan to explore utilizing glyph enhanced diffusion models for image-to-image translation \cite{DBLP:conf/acl/LanNMZZS24}.

\section*{Limitations}
While our methods enhance the visual text generation capability of the backbone models, several limitations still remain. First, our methods require to train the diffusion backbone model, which may be time consuming and expensive. Besides, our methods are unable to completely solve the issue of misspelling, ignoring and repeating words.

\section*{Ethics Statement}
This research paper rigorously addresses the ethical considerations associated with text-to-image models, ensuring that all methods used in this study are conducted responsibly and ethically. Our models are trained using open-source backbone models. To address concerns related to training data, we implement a strict filtering process to exclude inappropriate content, such as NSFW images and offensive visual text. The evaluation experiments are conducted using widely recognized public benchmarks, and participants involved in the user studies are systematically trained.

\section*{Acknowledgments}
The project is supported by National Key R\&D Program of China (No. 2022ZD0160501), National Natural Science Foundation of China (No. 62276219), Natural Science Foundation of Fujian Province of China (No. 2024J011001), and the Public Technology Service Platform Project of Xiamen (No.3502Z20231043). We also thank the reviewers for
their insightful comments.

\bibliography{acl_latex}

\appendix

\section{Dataset}
\label{sec:appendix_A}

\subsection{Data Collection}

To obtain suitable English training data, we initially survey academic datasets such as LAION-5B \cite{DBLP:conf/nips/SchuhmannBVGWCC22} and WuKong \cite{DBLP:journals/corr/abs-2202-06767}, and find that these academic datasets have issues with low image resolution, which significantly degrades the overall quality of the images when used for training. Therefore, we utilize our internal dataset and employ the high-precision OCR model PaddleOCR \cite{DBLP:journals/corr/abs-2009-09941} to filter data with texts. To get high quality captions including visual text information, we utilize a Multimodal Large Language Model (MLLM) and include the OCR results in the prompt to improve the accuracy of the generated captions. Following the above steps, we construct a English dataset of 240,000 high-aesthetic image-caption pairs.

\begin{figure}[b]
    \centering
    \includegraphics[width=0.48\textwidth]{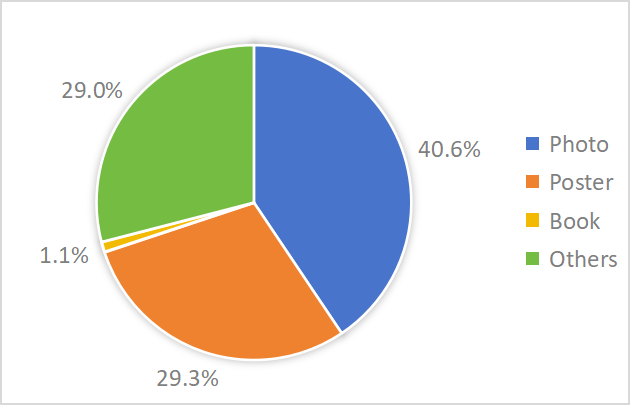}
    \caption{Statistics of data sources of our dataset.}
    \label{img.source}
\end{figure}

\begin{figure}[t]
    \centering
    \includegraphics[width=0.48\textwidth]{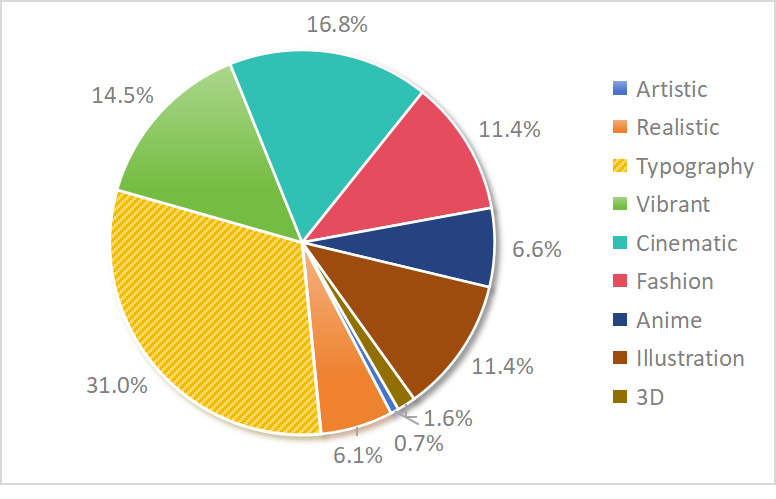}
    \caption{Statistics of data style of our dataset.}
    \label{img.style}
\end{figure}

Regarding Chinese data, we identify several issues with both web-crawled and academic datasets upon sampling: an excessive amount of text, overly complex glyphs, and text not occupying a prominent area in the images. These issues increase the difficulty of generating Chinese visual texts. Consequently, we explore constructing synthetic data using rendering and image-to-image models. We first select Chinese phrases that consists of two characters with no more than 10 strokes for each character \footnote{\url{https://github.com/thunlp/THUOCL/tree/master}}, and then apply manual deduplication to prevent overfitting to some characters, resulting in 255 phrases. We then design 10 templates to stipulate positional information and render the characters onto background images according to the templates. Finally, we apply image-to-image models to generate the backgrounds and conduct post-filtering with the OCR model to ensure that the aforementioned issues are avoided. However, due to the use of predefined rules, there are significant limitations in the diversity and overall aesthetic quality of the data, which impact the quality of the images generated by the model.

\begin{figure*}[t] 
    \centering \includegraphics[width=0.9\textwidth]{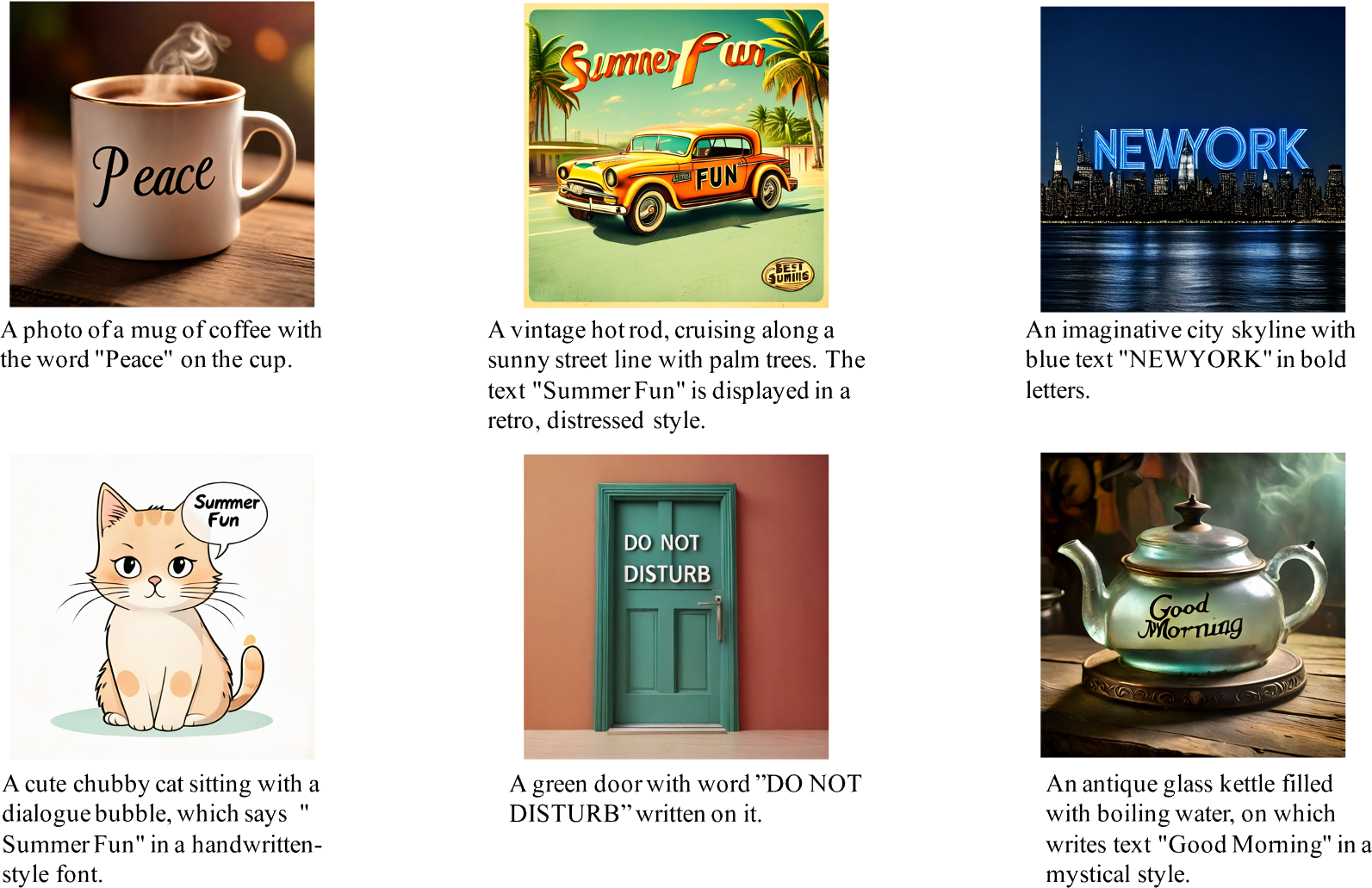}
    \caption{More visualizations of visual text generation results.}
    \label{Figure.more_showcase_text}
\end{figure*}

\subsection{OCR Filtering Rules}

We filter our collected data with the following criteria:

\begin{itemize}
\setlength{\itemsep}{0.1cm}
\setlength{\parsep}{2pt}
\setlength{\parskip}{2pt}
    \item \textbf{Height and width larger than 1024 for English dataset, and 512 for Chinese dataset.} We find that low-resolution sample has a negative impact for the training.
    \item \textbf{Area size for each visual text are more than 10\% of the whole image.} Visual texts being too small will increase the error rate of OCR recognition during training, thus introduce noise into data, and images with small texts often contain watermarks.
    \item \textbf{At least one detected text appears in the caption.} MLLMs would reject to describe when there is no visual text included in the image, we mark these images as invalid.
    \item \textbf{Text areas are at least 10\% away from border.} Texts too close to image boarder are more likely to be pruned when regulating images within a batch during training.
    \item \textbf{Number of texts should be no more than 5.} Samples that include too much texts typically have small areas for each text.
\end{itemize}

\begin{figure*}[t] 
    \centering \includegraphics[width=0.9\textwidth]{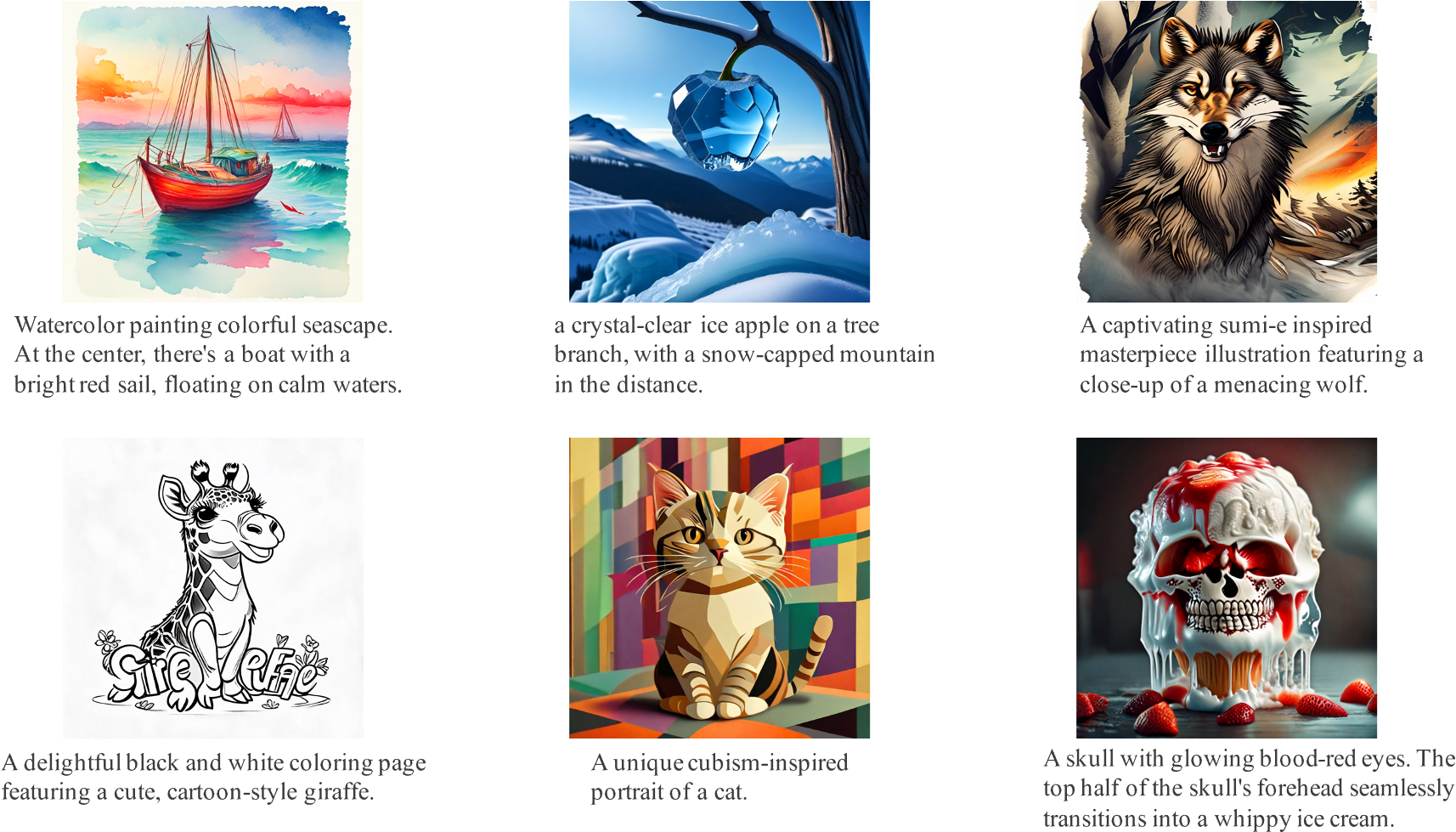}
    \caption{More visualizations of image generation without visual texts.}
    \label{Figure.more_showcase_wo_text}
\end{figure*}

\subsection{Data Statistics}

We further provide the statistics of our data source and style, as listed in Figure \ref{img.source} and \ref{img.style}.

\section{Participant Instruction}

\label{sec:appendix_B}

\textbf{Objective}: Evaluate the images based on the following four criteria. Select one or multiple images that you prefer; you can select all or none of them. Each sample will be evaluated by three raters.

\begin{itemize}
    \item \textbf{Text Accuracy}:
 
    \textit{Metric}: The visual texts should be correctly spelled and easily recognizable.
    The following errors should be considered in descending order of importance: neglecting or repeating words, misspelling words, generating words that are not requested.

    \textit{Further Consideration}:
    A repeated word may not be exactly the same as others. For example, if the prompt asks to generate the text 'apple', texts like 'aple', 'apple', 'appple' should all be considered as repeated words.

    \item \textbf{Text Aesthetics}:

    \textit{Metric}: The color of the visual texts should be coherent with the background.
    The font style of the visual texts should match the current scenario. For example, in paintings, the texts should be artistic; in posters, the texts should be eye-catching and occupy a prominent area.
    The positions of the visual texts should be reasonable.

    \textit{Further Consideration}:
    If the image does not contain any recognizable visual texts, it should be less preferred. The accuracy of the visual texts is not considered in this metric.

    \item \textbf{Semantic Relevance}:

    \textit{Metric}: The image should depict a scenario that matches the requirements of the user prompt.

    \textit{Further Consideration}:
    Note that the relevance between the visual texts and the user prompt is considered in text accuracy; the overall semantic relevance is more important.
    If the image contains noise from visual texts, it should be less preferred. For example, if the user prompt asks the image to contain the word 'apple' without needing to draw a real apple, images containing a real apple should be considered as bad cases.

    \item \textbf{Image Aesthetics}:

    \textit{Metric}: The image should be visually appealing.

    \textit{Further Consideration}:
    The aesthetics of the visual texts are less important than the background. 

\end{itemize}

\subsection{Scalability}

To investigate the scalability of our methods, we train three variants based on SD-XL, as detailed in Table \ref{figures:Scale}. The results demonstrate that as data scale and training steps increase, the model's performance consistently improves.


\begin{table}[]
\centering
\begin{small}
\setlength{\tabcolsep}{2.9mm}
\renewcommand\arraystretch{1.35}
{\begin{tabular*}{\linewidth}{@{}lcccc@{}} 
\toprule[1pt]
{\textbf{Variant}} & {\textbf{\makebox[0.075\textwidth]{CLIP score}}} &  {\textbf{{Pre.}}} & \textbf{{Rec.}} & \textbf{{F1}} \\ \hline
 


240K / 10K &29.168 & 0.348 & 0.283 & 0.312 \\
120K / 5K & 28.584 & 0.206 & 0.203 & 0.205  \\
48K / 2K & 28.075 & 0.127 & 0.173 & 0.146 \\

\bottomrule[1pt]
\end{tabular*}}
\caption{Quantitative results of scalability of our methods, A / B denotes the model is trained on A samples for B steps.}
\label{figures:Scale}
\end{small}
\vspace{-0.3cm}
\end{table}

\section{More Visualization Results}
\label{sec:appendix_C}

As depicted in Figure \ref{Figure.more_showcase_text} and Figure \ref{Figure.more_showcase_wo_text}, we showcase more visualizations results of our models on visual text generation task. Our model can generate visual appealing, style diverse, and legible visual text images, while maintaining basic capability to generate images without visual texts.


\begin{table}[]
\centering
\begin{small}
\setlength{\tabcolsep}{3.7mm}
\renewcommand\arraystretch{1.35}
\begin{tabular*}{\linewidth}{@{}lccc@{}} 
\toprule[1pt]
{\textbf{Method}} & {\textbf{{Pre.}}} & \textbf{{Rec.}} & \textbf{{F1}} \\ \hline

Ours & \textbf{0.598} & \textbf{0.613} & \textbf{0.606} \\
\hspace{1em} \(\Rightarrow\) BPE tokenization & 0.0 & 0.0 & 0.0 \\
\hspace{1em} w/o \(\mathcal{L}_{attn}\) & 0.578 & 0.604 & 0.591 \\
\hspace{1em} w/o \(\mathcal{L}_{loc}\) & 0.116 & 0.120 & 0.118  \\
\hspace{1em} w/o \(\mathcal{L}_{ocr}\) & 0.411 & 0.420 & 0.416 \\ 
\bottomrule[1pt]
\end{tabular*}
\caption{Ablation on the generation of Chinese texts. \(\Rightarrow\) BPE tokenization means using only BPE tokenization. \(\mathcal{L}_{loc}\), \(\mathcal{L}_{attn}\) and  \(\mathcal{L}_{ocr}\) denotes the local MSE loss , the attention alignment loss , and the OCR recognition loss, respectively}
\label{figures:AblationZH}
\end{small}
\end{table}

\section{Ablation Study for Chinese Text Generation}
\label{sec:appendix_D}
We further conduct ablation study for Chinese texts and compare the results with English texts.

(1) \textit{Char+BPE tokenization\(\Rightarrow\)BPE tokenization.} \quad We replace the mixture of character-level and BPE tokenization with BPE tokenization. While considering words as whole units achieves the best results for English texts, we find that utilizing character-level tokenization for Chinese texts yields the best performance. We indicate that this is because the glyph information for each Chinese character is more complex, and fewer characters are included in each phrase.

(2) \textit{w/o \(\mathcal{L}_{attn}\)}. \quad We remove the attention alignment loss \(\mathcal{L}_{attn}\) in this variant. From line 3, we observe a greater performance decline of OCR accuracy for English texts than Chinese texts without the attention alignment loss, indicating that English texts are more susceptible to cross-attention scores. We assume this is because more English words are included in each input image during training, making it harder to bind each visual text to its text token.

(3) \textit{w/o \(\mathcal{L}_{loc}\)}. \quad The local MSE loss \(\mathcal{L}_{loc}\) is not included in this variant. As shown in line 4, we observe a greater performance decline for Chinese texts when local MSE loss is not incorporated. This indicates that Chinese glyphs are harder to learn and should receive more attention during training.

(4) \textit{w/o \(\mathcal{L}_{ocr}\)}\quad We remove the OCR recognition loss \(\mathcal{L}_{ocr}\). As shown in line 5, similar to results for English texts, there is a significant performance decline when OCR recognition loss is not included,  the which indicates that the OCR recognition loss does have a positive effect for both English and Chinese texts.

\end{document}